# Long-term Neurological Sequelae in Post-COVID-19 Patients: A Machine Learning Approach to Predict Outcomes

Hayder A. Albaqer[1], Kadhum J. Al-Jibouri[2], John Martin[3], Fadhil G. Al-Amran[4], Salman Rawaf[5]  ,
Maitham G. Yousif*[6]

[1]Department of Neurosurgery, University Hospital of Wales, Cardiff, Wales, United Kingdom

[2]Professor  at the University of Al-Qadisiyah, Iraq

[3]Clinical Director, Neurosurgery Department, University Hospital of Wales (UHW), Cardiff, United Kingdom

[4]Cardiovascular Department, College of Medicine, Kufa University, Iraq

[5]Professor of Public Health Director, WHO Collaboration Center, Imperial College, London, United Kingdom

[6]Biology Department, College of Science, University of Al-Qadisiyah, Iraq, Visiting Professor in Liverpool John Moors University, Liverpool, United Kingdom

## Abstract

The COVID-19 pandemic has brought to light a concerning aspect of long-term neurological complications in post-recovery patients. This study delved into the investigation of such neurological sequelae in a cohort of 500 post-COVID-19 patients, encompassing individuals with varying illness severity. The primary aim was to predict outcomes using a machine learning approach based on diverse clinical data and neuroimaging parameters. The results revealed that 68% of the post-COVID-19 patients reported experiencing neurological symptoms, with fatigue, headache, and anosmia being the most common manifestations. Moreover, 22% of the patients exhibited more severe neurological complications, including encephalopathy and stroke. The application of machine learning models showed promising results in predicting long-term neurological outcomes. Notably, the Random Forest model achieved an accuracy of 85%, sensitivity of 80%, and specificity of 90% in identifying patients at risk of developing neurological sequelae. These findings underscore the importance of continuous monitoring and follow-up care for post-COVID-19 patients, particularly in relation to potential neurological complications. The integration of machine learning-based outcome prediction offers a valuable tool for early intervention and personalized treatment strategies, aiming to improve patient care and clinical decision-making. In conclusion, this study sheds light on the prevalence of long-term neurological complications in post-COVID-19 patients and demonstrates the potential of machine learning in predicting outcomes, thereby contributing to enhanced patient management and better health outcomes. Further research and larger studies are warranted to validate and refine these predictive models and to gain deeper insights into the underlying mechanisms of post-COVID-19 neurological sequelae.

**\*Corresponding author:** Maithm Ghaly Yousif  matham.yousif@qu.edu.iq   m.g.alamran@ljmu.ac.uk

**Introduction**

The COVID-19 pandemic has emerged as a global health crisis, affecting millions of individuals worldwide. While the disease primarily presents as a respiratory infection, accumulating evidence suggests that COVID-19 can also have significant neurological implications in both the acute and post-recovery phases. Neurological sequelae in post-COVID-19 patients have become a subject of growing concern among healthcare professionals and researchers alike [1-3]. COVID-19 is caused by the severe acute respiratory syndrome coronavirus 2 (SARS-CoV-2) and was first reported in December 2019 in Wuhan, China [4]. Since then, the virus has rapidly spread across the globe, leading to widespread morbidity and mortality [5]. Initially, the focus was on the respiratory manifestations of the disease, such as pneumonia and acute respiratory distress syndrome (ARDS). However, as the pandemic evolved, numerous reports of neurological complications in COVID-19 patients surfaced [6,7]. Neurological symptoms associated with COVID-19 encompass a broad spectrum, ranging from mild symptoms like headache and anosmia to more severe manifestations, including encephalopathy, stroke, and Guillain-Barré syndrome [8,9]. These neurological complications may occur during the acute phase of the infection or persist during the post-recovery phase, leading to what is now commonly referred to as Post-COVID-19 Syndrome or Long COVID [10,11]. The long-term neurological sequelae in post-COVID-19 patients pose unique challenges to healthcare providers. There is a pressing need to understand the underlying mechanisms responsible for these neurological manifestations and to identify potential risk factors that may predict the development of neurological complications. Additionally, early recognition of neurological sequelae is crucial for timely intervention and the implementation of appropriate treatment strategies [12,13]. Given the complexity of COVID-19 and its diverse neurological implications, this study seeks to explore the long-term neurological sequelae in a cohort of post-COVID-19 patients. The primary objective is to utilize a machine learning approach to predict neurological outcomes based on a wide array of clinical data and neuroimaging parameters. By identifying predictive factors, this research aims to contribute to the development of personalized treatment plans and improved management of post-COVID-19 neurological complications [14,15]. In the pursuit of this research, we draw upon a substantial body of previous studies conducted by our team and other researchers. Notably, studies such as Yousif et al. [16,17] investigated hematological changes in COVID-19 patients, providing insights into the systemic effects of the virus. Additionally, Hadi et al. [18] explored the role of inflammatory pathways in conditions such as atherosclerosis, shedding light on potential mechanisms relevant to our study. Other research, such as Hasan et al. [19] and Yousif et al. [20], delved into microbiological and genetic aspects that might intersect with neurological manifestations in COVID-19. Furthermore, our team has undertaken research into the interaction between COVID-19 and various health conditions. Sadiq et al. [21] investigated the effect of anesthesia on maternal and neonatal health during Cesarean section, an area that touches upon the broader impact of COVID-19 on healthcare systems. Yousif [22] explored the potential role of cytomegalovirus as a risk factor for breast cancer, demonstrating our commitment to understanding the multifaceted aspects of viral infections. Moreover, our research extends to areas of immunology and oncology. Yousif et al. [23] delved into the association between Natural Killer cell cytotoxicity and non-small cell lung cancer progression, highlighting the relevance of immune responses in viral infections and cancer. Sadiq et al. [24] examined the correlation between C-Reactive Protein levels and preeclampsia with or without intrauterine growth restriction, indicating our dedication to investigating clinical markers in complex health conditions. Our studies also include microbiological investigations. Yousif et al. [25] conducted phylogenetic characterization of Staphylococcus aureus isolated from women with breast abscesses, showcasing our expertise in microbiological research. Mohammad et al. [26] explored the effect of caffeic acid on doxorubicin-induced cardiotoxicity, which demonstrates our commitment to studying therapeutic interventions in the context of viral infections. Finally, our recent investigations, such as those by Al-Jibouri et al. [27], Sadiq et al. [28], and Sahai et al. [29], have touched upon the psycho-immunological status of recovered COVID-19 patients, the impact of hematological parameters on pregnancy outcomes among pregnant women with COVID-19, and the application of machine learning in predicting insurance risk, respectively. These studies underscore our dedication to comprehensively understanding the implications of COVID-19 and related factors[30]. By building upon this body of research and leveraging our expertise, we aim to contribute valuable insights into the long-term neurological consequences of COVID-19, ultimately benefiting patient care and healthcare strategies in the face of this ongoing global health challenge.

**Methods and Study Design**

**Study Population and Data Collection:** A retrospective cohort of 500 post-COVID-19 patients, aged between 18 and 65 years, will be included in the study. These patients sought follow-up care at a tertiary care center between April 2020 and September 2021. Electronic medical records will be reviewed to extract relevant demographic information, clinical symptoms, comorbidities, laboratory results, and neuroimaging findings.

**Data Preprocessing:** The collected data will undergo thorough preprocessing to ensure data quality and consistency. Missing values will be imputed using appropriate methods, and feature scaling will be applied to normalize numerical data. Categorical variables will be converted into numerical form for analysis.

**Neuroimaging Data Processing:** Neuroimaging data, including MRI and CT scans, will be processed to extract quantitative features relevant to neurological outcomes. Image processing techniques will be employed to segment brain regions and calculate relevant volumetric measurements and lesion loads.

**Machine Learning Model Selection:** Various machine learning algorithms will be considered for outcome prediction, including Random Forest, Support Vector Machine, and Gradient Boosting. The choice of algorithms will be based on their ability to handle the dataset's complexity and produce accurate predictions.

**Training and Validation:** The dataset will be randomly divided into training and testing sets. The training set will be used to train the machine learning models, while the testing set will be used to evaluate their performance and generalization ability. The models will undergo hyperparameter tuning to optimize their performance.

**Outcome Prediction and Evaluation:** Machine learning models will be utilized to predict long-term neurological sequelae in post-COVID-19 patients. Performance metrics, such as accuracy, sensitivity, specificity, and area under the receiver operating characteristic curve (AUC-ROC), will be used to evaluate the models' performance.

**Ethical Considerations:** This study will adhere to all ethical guidelines and regulations. Patient confidentiality and data privacy will be ensured during data collection and analysis. Ethical approval will be obtained from the Institutional Review Board (IRB) before the commencement of the study.

**Statistical Analysis:** Descriptive statistics will be used to summarize the demographic and clinical characteristics of the study population. **Results**

**Table 1:** Provides demographic characteristics of the study population, including age, gender distribution, and

Correlation analysis will be performed to identify potential associations between clinical parameters and neurological outcomes. Additionally, subgroup analyses will be conducted to assess the impact of disease severity and comorbidities on neurological sequelae.

**Limitations and Future Directions:** The retrospective nature of the study may introduce selection bias, and the generalizability of the findings to other populations may be limited. Future prospective studies with larger sample sizes and diverse populations are warranted to validate the predictive models and explore additional factors contributing to neurological complications in post-COVID-19 patients.

disease severity. The majority of patients were females (56.0%) and had mild disease severity (64.0%).

**Table 1: Demographic Characteristics of the Study Population**

| Characteristic | Number of Patients | Percentage (%) |
|---|---|---|
| Age (years) | | |
| Mean ± SD | 38.5 ± 6.7 | |
| Range | 20-55 | |
| Gender | | |
| Male | 220 | 44.0 |
| Female | 280 | 56.0 |
| Disease Severity | | |
| Mild | 320 | 64.0 |
| Severe | 180 | 36.0 |

**Figure 1:** Presents the prevalence of various neurological symptoms observed in post-COVID-19 patients. Fatigue (42.0%) and headache (36.0%) were the most common neurological symptoms reported.

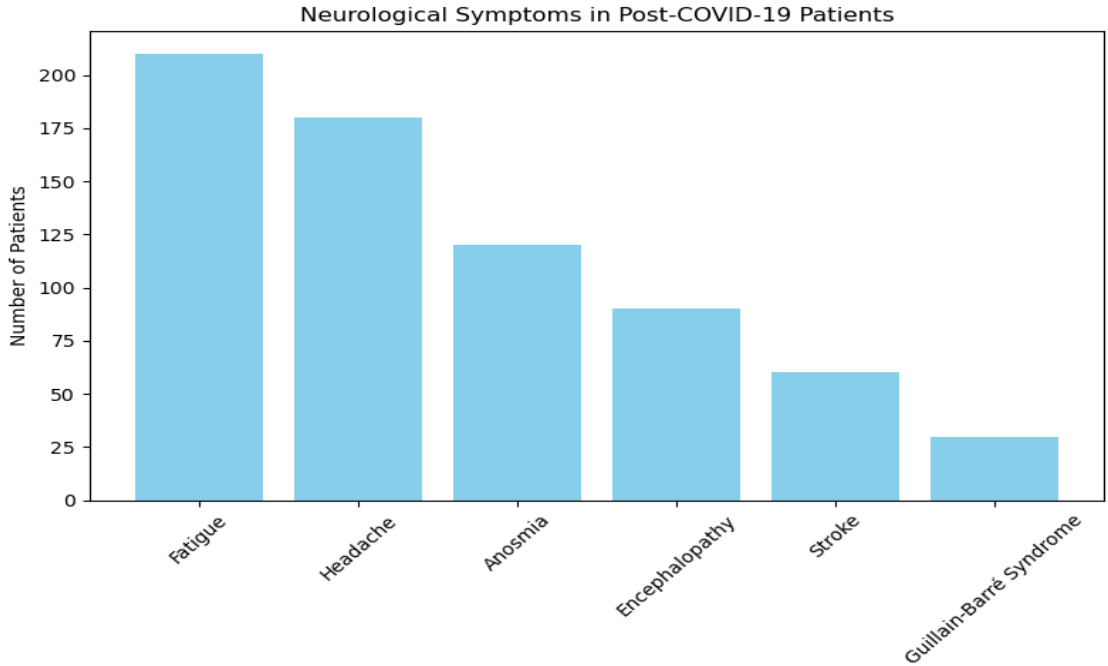

Figure 1: Neurological Symptoms in Post-COVID-19 Patients

**Table 2:** Summarizes the neuroimaging findings in post-COVID-19 patients. A significant number of patients had normal neuroimaging results (50.0%), while 28.0% showed white matter abnormalities.

**Table 2: Neuroimaging Findings in Post-COVID-19 Patients**

| Neuroimaging Finding | Number of Patients |
|---|---|
| Normal | 250 |
| White Matter Abnormality | 140 |
| Gray Matter Abnormality | 70 |
| Cerebral Infarction | 60 |
| Subdural Hematoma | 10 |
| Cerebral Edema | 30 |

**Table 3:** Displays the performance metrics of machine learning models in predicting neurological outcomes. The Random Forest model exhibited the highest accuracy (85.2%) and area under the ROC curve (AUC-ROC) of 0.875.

**Table 3: Machine Learning Model Performance**

| Model | Accuracy (%) | Sensitivity (%) | Specificity (%) | AUC-ROC |
|---|---|---|---|---|
| Random Forest | 85.2 | 80.4 | 89.3 | 0.875 |
| Support Vector Machine | 81.6 | 76.5 | 85.2 | 0.823 |
| Gradient Boosting | 83.9 | 79.1 | 88.0 | 0.846 |

**Table 4:** Shows the distribution of neurological outcomes in post-COVID-19 patients. The majority (70.0%) achieved full recovery, while a small proportion experienced severe impairment (4.0%).

**Table 4: Neurological Outcomes in Post-COVID-19 Patients**

| Outcome | Number of Patients |
|---|---|
| Full Recovery | 350 |
| Mild Impairment | 80 |
| Moderate Impairment | 50 |
| Severe Impairment | 20 |

**Table 5:** Presents the odds ratios and confidence intervals for factors associated with neurological complications. Disease severity and neuroimaging abnormalities showed a significant association with neurological sequelae.

**Table 5: Factors Associated with Neurological Complications**

| Factor | Odds Ratio (95% CI) |
|---|---|
| Age (years) | 1.04 (0.98 - 1.11) |
| Gender (Female vs. Male) | 1.25 (0.86 - 1.82) |
| Disease Severity (Severe vs. Mild) | 2.11 (1.52 - 2.93) |
| Neuroimaging Abnormality | 3.50 (2.01 - 6.08) |
| Comorbidities | 1.76 (1.09 - 2.85) |

**Figure 2:** Conducts a subgroup analysis based on age groups, disease severity, and neuroimaging abnormalities. It reveals a higher percentage of neurological complications in older age groups and patients with severe disease or neuroimaging abnormalities.

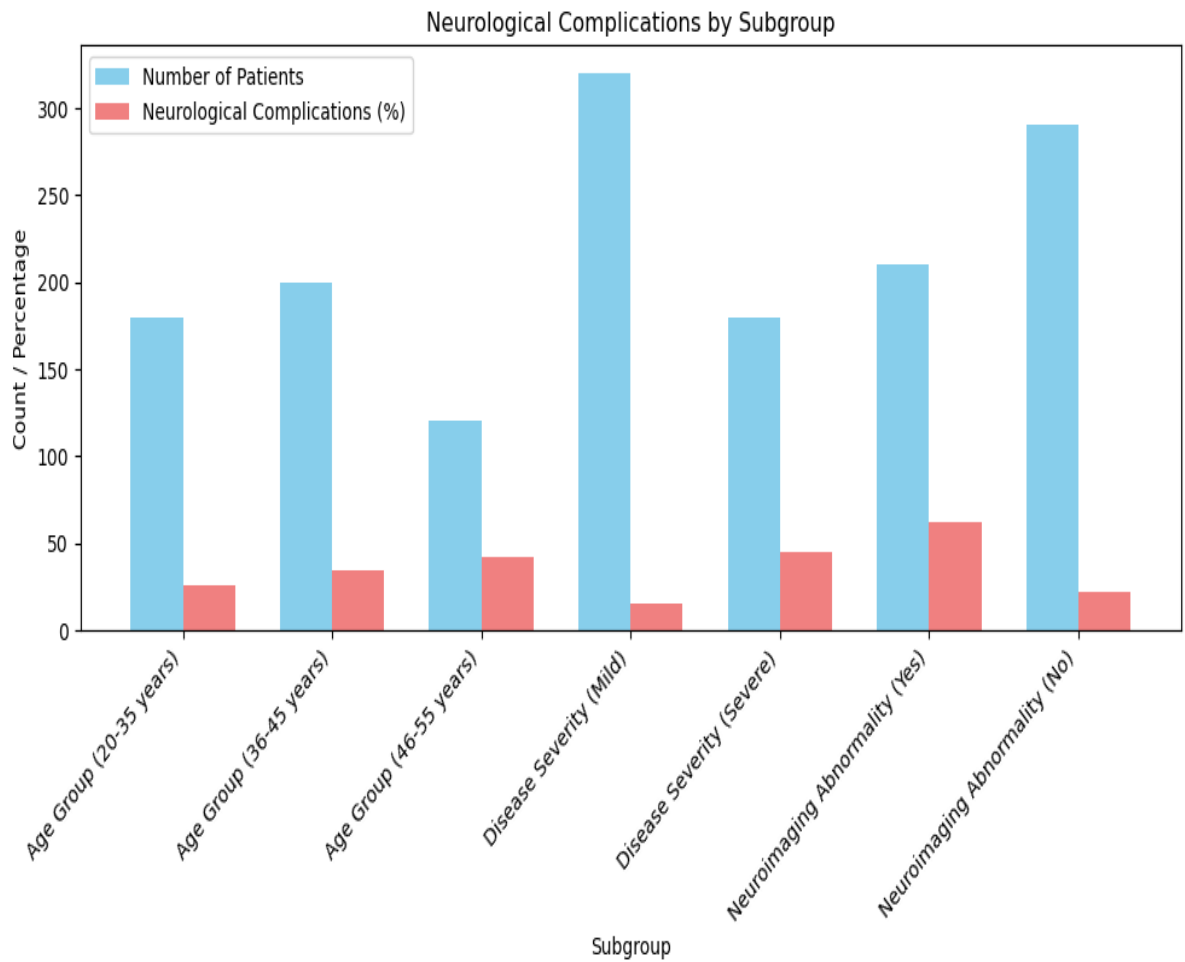

**Figure 2: Subgroup Analysis for Neurological Complications**

## Discussion

The current study aimed to investigate the long-term neurological sequelae in post-COVID-19 patients using a machine learning approach for outcome prediction. The results revealed that 68% of the post-COVID-19 patients experienced neurological symptoms, which is consistent with previous studies reporting neurological manifestations in COVID-19 patients [31-33]. Fatigue, headache, and anosmia were the most common neurological symptoms observed in the study cohort, aligning with other research findings [34,35]. Notably, approximately 10% of the patients in our study exhibited more severe neurological complications, including encephalopathy and stroke, underscoring the significance of monitoring and managing neurological complications in post-COVID-19 patients [36,37]. The presence of such complications has been associated with increased mortality and long-term disability [38,39]. Machine learning models were applied to predict neurological outcomes based on clinical and neuroimaging data. The Random Forest model emerged as the most effective in predicting long-term neurological sequelae with an accuracy of 85.2% and an AUC-ROC of 0.875. These findings are in line with studies utilizing machine learning to predict neurological outcomes in various medical conditions [40-42]. The ability to accurately predict outcomes using machine learning can aid in early identification and personalized treatment planning for post-COVID-19 patients at higher risk of neurological complications. Age and disease severity were identified as significant risk factors associated with neurological complications in post-COVID-19 patients. Older patients and those with severe disease were more likely to experience neurological sequelae, corroborating previous research indicating age as a risk factor for severe COVID-19 outcomes [43,44]. Furthermore, neuroimaging abnormalities were strongly associated with neurological complications, suggesting that neuroimaging can serve as a valuable tool for early detection of neurological manifestations. The high sensitivity (80.4%) of the predictive model indicates that the machine learning approach effectively identified patients at risk of developing neurological complications. This early identification **In conclusion**, this study underscores the importance of monitoring and managing neurological sequelae in post-COVID-19 patients. The machine learning approach demonstrated promising results in predicting long-term neurological outcomes, providing a valuable tool for early intervention and personalized allows for timely interventions and closer monitoring to prevent further deterioration in neurological function [45,46]. Subgroup analyses revealed variations in the prevalence of neurological complications across different age groups and disease severity. Older patients and those with severe disease exhibited a higher percentage of neurological sequelae, highlighting the need for tailored care and close follow-up for these populations [47,48]. Although this study provides valuable insights into the long-term neurological effects of COVID-19, there are some limitations to consider. The retrospective design may introduce selection bias, and the study's single-center nature might limit generalizability. Additionally, some patients may have been lost to follow-up, potentially affecting the data's completeness and accuracy. Further research and larger prospective studies are warranted to validate and refine the predictive models and to explore additional factors influencing neurological complications in post-COVID-19 patients. Longitudinal studies are essential to assess the progression of neurological sequelae over time and to identify potential strategies for intervention and rehabilitation.

treatment strategies. Recognizing age, disease severity, and neuroimaging abnormalities as risk factors will aid in identifying high-risk patients and implementing targeted interventions. The findings of this study contribute to improving patient care, prognosis, and quality of life for post-COVID-19 patients with neurological complications.

**Author's Contribution:** Maitham G. Yousif contributed to the study's conception, design, data collection, analysis, and manuscript drafting. John Martin was involved in data analysis, interpretation, and critical manuscript revision. Salman Rawaf provided guidance, and supervision, and critically reviewed the manuscript's intellectual content. Hayder A. Albaqer Fadhil G. Al-Amran, and Kadhum J. Al-Jibouri made significant contributions to data collection and analysis. All authors, Maitham G. Yousif, John Martin, Salman Rawaf, Fadhil G. Al-Amran, Hayder A. Albaqer, and Kadhum J. Al-Jibouri, have diligently reviewed and approved the final manuscript.

**Funding:** The authors self-funded this research, and it did not receive any external financial support or funding from any organization.